\documentclass{article}
\usepackage{arxiv}

\usepackage{amsmath}
\usepackage{amssymb}
\usepackage{amsfonts}       

\usepackage[utf8]{inputenc} 
\DeclareUnicodeCharacter{2212}{-}

\usepackage[T1]{fontenc}    
\usepackage{url}            

\usepackage{multirow}
\usepackage{mathtools}
\usepackage{booktabs}       
\usepackage{caption}
\usepackage{subcaption}

\usepackage[linesnumbered, ruled, vlined]{algorithm2e}
\usepackage{array}

\usepackage{graphicx}
\usepackage{hyperref}
\usepackage[gen]{eurosym}

\usepackage{nicefrac}       
\usepackage{microtype}      

\newcommand{\bigCI}{\mathrel{\text{\scalebox{1.07}{$\perp\mkern-10mu\perp$}}}} 

\title{Affordable Uplift: Supervised Randomization in Controlled Experiments}

\begin{document}

\maketitle

\vspace{0.5cm}
\begin{center}
    \textbf{Johannes Haupt, Daniel Jacob, Robin M. Gubela, Stefan Lessmann}\\
    Humboldt-Universit\"at zu Berlin\\
    \{johannes.haupt, daniel.jacob, robin.gubela, stefan.lessmann\}@hu-berlin.de
\end{center}
\vspace{0.5cm}

\begin{abstract}

Customer scoring models are the core of scalable direct marketing. Uplift models provide an estimate of the incremental benefit from a treatment that is used for operational decision-making. Training and monitoring of uplift models require experimental data. However, the collection of data under randomized treatment assignment is costly, since random targeting deviates from an established targeting policy. 
To increase the cost-efficiency of experimentation and facilitate frequent data collection and model training, we introduce \textit{supervised randomization}. It is a novel approach that integrates existing scoring models into randomized trials to target relevant customers, while ensuring consistent estimates of treatment effects through correction for active sample selection. 
An empirical Monte Carlo study shows that data collection under supervised randomization is cost-efficient, while downstream uplift models perform competitively. 

\end{abstract}

\section{Introduction}

Direct marketing plays a key role in consumer markets. The continuous growth of e-commerce, accounting for 1.8 trillion Euros globally in 2019 \cite{statista2019ecommerce}, is accompanied by simultaneous growth in online and email advertising. Spending on traditional print advertising like catalog marketing has shown a similar growth \cite{statista2017advertising}. At the core of scalable direct marketing, campaign analysts employ models to predict future customer behavior and target responsive clients \cite{olson2012direct}. 

For example, a decision tree could be trained to predict the probability for a customer to purchase in the next week based on known characteristics. The expected behavior of the customer could then be used to inform operational decision-making in that customers with a probability below average are targeted with an incentive. However, the predictive model is agnostic to the marketing policy, the overall effectiveness of the marketing action and the effect of the marketing action on individual customers. Outcome models provide an estimate of customer behavior, but fail to provide an estimate of the potential change in customer behavior, which is the goal of marketing intervention. 

A growing research stream advocates that the decision which customers to target should be addressed directly through causal inference in the form of uplift models \cite{gubela2017revenue, devriendt2018literature}. Instead of predicting customer behavior, uplift models estimate the causal effect of a marketing action on an individual customer given their characteristics. In the above example, an uplift tree could be trained to predict the increase in probability for a customer to conduct a purchase in the next week if a catalog was sent. 
Uplift models thus provide an estimate of the incremental benefit from the marketing treatment, which can explicitly be used as a direct criterion for operational decisions by comparing it to the incremental cost. Conceptually, uplift models align with the actual decision problem of choosing the action with the highest incremental gain for each customer. 

Uplift models are trained on experimental data and estimate the treatment effect by comparing the observed behavior of a group of individuals who have received the treatment, the treatment group, and a distinct group of individuals who have not received the treatment, the control group. Similarly, experimental data is required to evaluate the performance of uplift models \cite{radcliffe2007using}. In contrast, non-causal models of customer behavior are trained and evaluated on customers of which all or none have received the treatment. Collecting experimental data in randomized experiments is well established in practice in the form of A/B tests. Although used to evaluate the gross benefit of campaigns, A/B tests are not commonly used for uplift modeling to estimate individualized treatment effects \cite{ascarza2018RetentionFutilityTargeting}. 

During experiments, random assignment of individuals to either the treatment or control group is crucial to train unbiased uplift models. However, data collection through randomized experiments is costly, since random targeting withholds marketing spending on some customers that would be targeted under the established targeting policy and applies spending on customers that would otherwise not be targeted. The deployment of uplift models exacerbates data collection costs since decision support systems typically require continuous or frequent evaluation and occasional retraining on recent observations, which in turn require fresh experimental data. 

We propose a novel approach for the collection of experimental data for uplift modeling based on the combination of cost-optimized randomization at the time of data collection and selection bias correction during model building, which we refer to as \textit{supervised randomization}. In a nutshell, supervised randomization introduces a stochastic component to the existing targeting model and extends the standard experimental design with full randomization by considering customers that are rejected by the targeting policy. \\
Our contribution is two-fold.
First, we show that supervised randomization can be used to integrate existing scoring models into randomized trials. The integration of existing scoring models into group assignment increases the cost-efficiency of experimentation and facilitates continuous data collection during regular business operation. Continuous data collection is critical for non-disruptive experimentation, monitoring the performance of uplift models under deployment and recurring model training. Facilitating model training and monitoring has the additional benefit to improve the acceptance of causal models by management and stakeholders. \\
Second, we introduce inverse probability weighting and doubly robust estimation as methods to control for biased treatment assignment to the uplift literature. Uplift models have so far relied on the assumption of data collected under full or imbalanced randomization in randomized controlled trials. We show that recent advances in the econometrics literature extend the applicability of uplift models to cases with non-standard treatment assignment. The bias-corrected uplift models are shown to perform competitively on simulated data.

\section{Background}
Consider a marketing action applied to an individual user $i$ as a treatment intended to change an observed outcome $Y_i$. Let $D_i \in {0,1}$ be an indicator if the individual has been treated and denote the outcome with and without treatment as $Y_i^1$ and $Y_i^0$, respectively. Then the individual treatment effect is the incremental gain caused by a marketing action $Y_i^1 - Y_i^0$. Because a customer either does or does not receive the marketing action, the actual treatment effect is not observable. We can, however, estimate the treatment effect on the population or on the individual level. We denote the average campaign uplift as average treatment effect (ATE) and the customer-level uplift  $ \tau = E[Y_i^1 - Y_i^0|x = X_i]$ as individualized treatment effect (ITE) \cite{powers2018methods, knaus2018machine}, sometimes alternatively denoted conditional average treatment effect (CATE) in the econometrics literature. Furthermore, we refer to a model used to estimate the outcome $Y_i$ as outcome model and a model used to estimate the treatment effect $\tau$ as a causal model, as a more general alternative to the term uplift model. The operational decision problem posed in uplift modeling is to decide if an individual customer should receive the marketing treatment. The decision is automated through the targeting policy, a mapping from the estimated ITE to the binary decision of whether to treat the individual. 

Three assumptions are needed for causal inference following the potential outcome theorem \cite{rosenbaum1983central}. First, the \textit{Stable Unit Treatment Value Assumption (SUTVA)} guarantees that the potential outcome of a customer is unaffected by changes in the treatment assignment of other customers. This assumption may be violated when treatment effects propagate through the social network of customers \cite{Ascarza2017target}. In settings of low value or low involvement products, research on treatment effects in marketing typically assumes that no interaction takes place \cite{hitsch2018heterogeneous} or that the indirect effect of treatment on other customers is at least substantially smaller than the direct effect of the treatment \cite{imbens2009RecentDevelopmentsEconometrics}.  

The second assumption is \textit{conditional unconfoundedness}, i.e. the independence between the potential outcomes and the treatment assignment given the observed covariates ($X$) \cite{rosenbaum1983central}. 

\begin{align}
\left(Y_{i}^{1}, Y_{i}^{0}\right) \bigCI   D_{i}|X_{i} = x. \label{eq:unconfoundedness}
\end{align}
The third assumption, called \textit{overlap}, guarantees that for all $x \in supp(X_i)$  the probability to receive treatment $e(x) = \operatorname{P}(D=1|X_i=x)$ is bounded away from 0 and 1:
\begin{align}
0 < e(x) < 1. \label{eq:overlap}
\end{align}
When the treatment assignment process is under the control of the experiments as in the customer targeting setting, conditional independence and overlap can be ensured by design through fully randomizing treatment assignment with treatment probability $e(x) = e \in (0;1)$. Randomized experiments assign individuals at random to one of at least two conditions, where each condition entails a specific treatment. In controlled experiments, one condition is the control condition in which individuals receive no treatment. In combination, randomized controlled trials (RCT) are the gold standard of data collection for causal inference. We refer to uniform treatment assignment as \textit{full randomization}. Supervised randomization provides a framework that preserves the advantage of the randomized experimental design but allows some control over the probability of treatment assignment on the individual level. 

\section{Literature Review}\label{Section:Literature}

The unbiased training of causal models and targeting policies requires data that fulfills the assumptions of the potential outcome theorem. In addition, the unbiased evaluation of causal models and policies also requires experimental data and metrics developed for counterfactual prediction \cite{radcliffe2007using, hitsch2018heterogeneous}. Violation of the unconfoundedness (Eq. \ref{eq:unconfoundedness}) and overlap assumptions (Eq. \ref{eq:overlap}) in observational studies cannot be substituted by collecting more data in the form of more covariates or more observations \cite{gordon2019ComparisonApproachesAdvertising}. Randomized experiments are thus considered a prerequisite to uplift modeling. However, the design and costs of RCT are often not discussed in the literature. We aim to fill this gap by proposing a more efficient design for randomized experiments. We first summarize recent developments in causal machine learning, related research on efficient experimental design and methods to correct for treatment assignment in observational studies. 

Causal machine learning methods can be divided into direct and indirect approaches. Direct estimation algorithms construct a feasible loss to estimate a model for the ITE. Indirect approaches model the expected customer response conditional on the treatment group and estimate the ITE as the difference between expected responses. This study employs a robust, indirect two-model logistic regression and a state-of-the-art, direct causal forest for the empirical comparison and provides a discussion of these models below. For an in-depth discussion and benchmark of recent methods for ITE estimation see \cite{powers2018methods,knaus2018machine,kunzel2019MetalearnersEstimatingHeterogeneousa}.

Indirect approaches estimate the treatment effect via estimating the response with and without treatment using common statistical learners. The two-model approach \cite{radcliffe2007using}, or K-model approach in settings with more than one treatment, estimates a separate model for the outcome in the treatment group and control group data and estimates the ITE as the difference between the predicted outcomes. The two-model approach is flexible with regard to the underlying outcome models. K-nearest neighbors learners \cite{gubela2019conversion} and deep neural networks \cite{farrell2018DeepNeuralNetworks} have demonstrated promising model performance in the two-model framework. While recent research advocates discretizing the outcome variable to use classification models in continuous settings \cite{gubela2017revenue}, the two-model approach extends naturally to both categorical and continuous outcomes. This facilitates the use of classification and regression models to forecast, for example, purchase completion or customer spending, respectively.

A number of well-known machine learning algorithms have been extended to estimate the ITE directly without the need to model the customer response \cite{lo2002true, zaniewicz2013support}. Note that the average treatment effect within a subgroup provides a useful estimate of the treatment effect for individuals within that subgroup. Hence, algorithms that split the data into groups to calculate estimates on the subset are inherently applicable to causal modeling and modifications of the k-nearest neighbor estimator \cite{hitsch2018heterogeneous} and tree-based models \cite{rzepakowski2012decision,athey2016recursive} have been applied to estimate individualized treatment effects. 
Causal tree models modify the Classification and Regression Tree by a splitting criterion maximizing the expected variance in treatment effects between leaves \cite{athey2016recursive}. Within each terminal node conditional on the covariate splitting, the conditional average treatment effect can be estimated and provides an ITE for the observations falling into that node. \\
Causal trees can be combined into ensembles through bagging or boosting. \cite{powers2018methods} propose a gradient-boosted ensemble of causal trees and an algorithm using multivariate adaptive regression splines. Causal forests are similarly flexible models and have been shown to be consistent and asymptotically normal for a fixed covariate space \cite{athey2019GeneralizedRandomForests}.

Both direct and indirect approaches to ITE estimation share the need for experimental data. The collection of experimental data has not been explicitly explored in the uplift literature. However, concerns over the organizational difficulty and the opportunity cost of running randomized controlled trials have led to research on the optimal use of available data and efficient experimental design in related fields. \\
A popular strategy for the evaluation of multiple targeting policies is to avoid experimentation for each candidate policy and instead to estimate each policy's performance using one existing, fully randomized experiment. The cost-efficient evaluation is possible through extrapolation from observations where the policy recommendation matches the observed random treatment assignment, weighted to match the actual population \cite{swaminathan2015BatchLearningLogged, athey2017EfficientPolicyLearning, hitsch2018heterogeneous}. 
This evaluation strategy requires existing experimental data, while our goal is to decrease the cost of collecting experimental data through efficient randomization. Approaches to efficient evaluation and efficient randomization are therefore complementary.

{
\begin{table}[ht]
\caption{Randomized treatment data in marketing}
\label{tab:uplift}
\centering
\begin{tabular}{lllrr}
\toprule
Application & Source &  Access & Observations & \begin{tabular}[c]{@{}r@{}}Treatment /\\ Control Ratio\end{tabular} \\ 
\midrule
Direct mail in office supplies & \cite{kane2014mining} & Closed & 460,000 & 17:1 \\ 
Mail promotion & \cite{hansotia2002direct} & Closed & 550,000 & 10:1 \\ 
Cross-selling mail in insurance & \cite{guelman2014optimal} & Closed & 34,370 & 9:1 \\ 
MSN subscription & \cite{chickering2000decision} & Closed & 110,000 & 9:1 \\ 
Criteo online advertising campaign & \cite{diemert2018large} & Open & 29,105,905 & 7:1 \\
Direct mail in financial services & \cite{kane2014mining} & Closed & 1,144,000 & 5:1 \\ 
Simulation study & \cite{lo2002true} & Closed & 100,000 & 4:1 \\ 
Customer retention mail in insurance & \cite{guelman2015uplift} & Closed & 11,968 & 2:1 \\ 
Catalog marketing & \cite{hitsch2018heterogeneous} & Closed & 441,000 & 2:1 \\ 
E-mail promotion in merchandising & \cite{hillstrom2008mine} & Open & 64,000 & 2:1 \\ 
E-mail promotion in holiday marketing & \cite{hansotia2002incremental} & Closed & 282,277 & 1:1 \\ 
\bottomrule
\end{tabular}
\end{table}
}

The experimental design of previous studies indicates awareness of data collection costs. Table \ref{tab:uplift} shows the marketing goal, data accessibility, number of observations and the imbalance between treatment and control group sizes of experimental campaigns in customer targeting applications. The large number of observations in recent studies is unsurprising since common technologies in e-commerce settings (e.g., web cookies) facilitate the collection of large data volumes of customer interactions in online shops. However, large-scale experimentation imposes substantial costs by randomly withholding profitable treatment for a sizeable control group. We reason that large experiment sizes indicate that companies perceive potential gains from causal modeling and are willing to collect data on a sufficient scale. Since the costs of experimentation are a result of the randomization of treatment assignment, we propose that supervised randomization can lead to cost reductions that are economically relevant in practice given the scale of experimentation. 
The savings potential increases with the targeting cost and will thus be most effective for catalog or telephone marketing, where resource-intensive treatments drive cost, and in customer churn management, where targeting customers may increase awareness of contract expiration and induce churn in otherwise passive customers.\\
We further observe that 10 of 11 datasets show a substantial difference in size between the treatment and control group, which we denote \textit{imbalanced full randomization}. The imbalance implies that companies assign customers to the treatment group with probabilities 2-17 times higher than assignment to the control group. The observed imbalanced experimental design is more efficient than equal assignment to treatment and control group when the marketing action is expected to be profitable on average and treatment is the dominant targeting strategy. Companies are thus conducting active cost management of random experiments based on an assessment of overall treatment effectiveness. Our approach follows the same motivation, but extends cost management to the individual level based on an assessment of the individual treatment effectiveness. 

The design of randomization on the individual level is more thoroughly discussed in the medical literature \cite{schulz2002generation}. On the one hand, administering a pharmaceutical to a random patient can induce severe health issues, so randomized trials pose a risk for patient health. On the other hand, new treatment may prove to be a substantial improvement over comparative options, so that withholding treatment can be seen as suboptimal care. The latter concern for optimal treatment of patients has motivated research on adaptive randomization procedures, where patients are more likely to be assigned to treatment for which positive outcomes have been previously observed over the course of the study \cite{lachin1988randomization, rosenberger1993use}. Response-adaptive randomization in medical trials differs from our approach in that we use a scoring model to adjust treatment probability conditional on customer characteristics rather than observing treatment outcomes during the trial.

The trade-off between collecting more data to improve the scoring model and applying an existing treatment policy deterministically corresponds to the exploration-exploitation problem in reinforcement learning and multi-armed bandit approaches. 
Supervised randomization is related to the $\varepsilon$-greedy algorithm \cite{schwartz2017CustomerAcquisitionDisplay}, extended by heterogeneous exploration probabilities $\varepsilon_i=1-e(x)$. In comparison to upper confidence bound sampling or Thompson sampling \cite{schwartz2017CustomerAcquisitionDisplay} which favor exploration of uncertain predictions, supervised randomization favors exploration close to the decision boundary of the policy and facilitates straightforward logging of the true treatment probability. \\
This study considers supervised randomization for continuous evaluation and periodical updating of treatment effect models. We do not adapt the scoring model and conditional treatment probabilities during the duration of the experiment as opposed to online learning of the treatment effect model under reinforcement learning. We leave a more in-depth comparison for future research.

Supervised randomization introduces dependency between the covariates and treatment assignment in the data as a side effect of adjusting treatment probabilities on the individual level. This violates the conditional independence assumption and without correction would lead to biased treatment estimates known as selection bias. An intuitive interpretation is that selection bias is due to the covariates being non-identically distributed between the treatment and control group because group assignment is itself based on the observed covariates. 
Statistical analysis of the average or individualized treatment effect on data that violates the unconfoundedness assumption thus requires correction for the effect of the covariates on the individual probability to receive treatment.  To the best of our knowledge, methods to systematically correct for selection bias have not yet been studied in the uplift community. Instead, research on uplift modeling assumes the feasibility of randomized control trials where there is no such selection bias by design, but ignores the associated costs of data collection in practice. \\
However, selection bias corrections are well-understood and common for research using observational data, where random treatment assignment is not ethical or feasible. The most common technique for selection bias correction are the inverse probability weighting estimator \cite{horvitz1952generalization} (IPW) and its extension to the doubly robust (DR) estimator \cite{robins1994estimation}. Selection bias correction has recently been integrated into popular ITE estimators, which are applied in the observational studies prevalent in economics research \cite{athey2019GeneralizedRandomForests, kunzel2019MetalearnersEstimatingHeterogeneousa}. 

Within the field of information systems, the IPW correction is used in observational studies \cite{caliendo2012cost} and research on recommender systems and reinforcement learning. In recommendation settings, explicit and some forms of implicit feedback can be understood as the outcome of a non-randomized experiment, where users evaluate a subset of items that they select based on their preferences. The customer feedback can be corrected by weighting the feedback by the probability of a customer to interact with an item before rating it \cite{schnabel2016recommendations}. \\
Reinforcement learning research makes regular use of existing log data, which is cost-efficient to collect and available at scale, but not fully randomized. Under some conditions, unbiased training or evaluation of a learning algorithm is possible using IPW to correct for the treatment policy at the time of data collection \cite{swaminathan2015BatchLearningLogged}. Interestingly, if treatment under the existing policy is stochastic with a known probability and the treatment probability is logged, the resulting process can be seen as an online version of the supervised randomization process.

For observational data, the treatment probability used to correct for the selection bias is unknown and IPW thus follows a two-step process. In the first step, the treatment probabilities used for IPW correction are estimated from the observed data. In the second step, the treatment probability estimates are used to correct subsequent estimates of treatment effects. The approach proposed here is substantially different from the common applications of IPW in the first step. Under supervised randomization, the true treatment probability for each observation is actively controlled and thus known, eliminating the need for estimation and the potential for estimation error through unobserved variable bias or model misspecification. 

\section{Efficiently Randomized Experimental Design}

Within this study, we take a holistic view of the causal modeling process emphasizing the interaction between data collection and model building. Data collection through a randomized controlled trial is a necessary part of the causal modeling process. To collect RCT data, the established targeting policy is temporarily replaced by randomized treatment assignment. However, the replacement of an efficient targeting strategy by a random assignment has negative side effects in practice, even when restricted to a subset of customers. First, randomized treatment assignment carries opportunity costs resulting from targeting the wrong customers. Compared to an existing effective targeting strategy, profitable customers are less likely to be targeted while less profitable customers are more likely to be targeted. Second, customers may misconceive data collection periods as a decrease in service or advertising quality. Since customers are not informed about the temporary replacement of the targeting model, they will attribute the random treatment assignment to the targeting efforts of the company. 

Instead of replacing the established targeting policy with randomized treatment allocation, we propose to introduce a stochastic component to the existing targeting model as \textit{supervised randomization}. Under supervised randomization, treatment assignment is largely driven by the effective targeting model but sufficiently randomized to allow the estimation and evaluation of causal models. Embedding the existing targeting model into the experimental design has three merits. \\
First, supervised randomization increases the return on treatment during experimentation when compared to full randomization. Supervised randomization allows us to actively decrease the cost of running randomization experiments by treating profitable customers identified by the targeting model with a higher probability than customers identified as less profitable by the scoring model. \\
Second, supervised randomization with conservative propensity mapping could facilitate continuous experimentation. Since both training and evaluation of causal models require experimental data, regular repetition of experimental data collection is necessary when causal models are deployed. Continuous experimentation could reduce variability in service from the perspective of the customer and streamline data collection for the company by avoiding interference with operation to run experiments and reducing the need to justify and approve the need of data collection at intervals.
These goals are shared with reinforcement or bandit learning with the important difference that our approach facilitates model estimation through standard machine learning or uplift methodology. 

\subsection{Supervised randomization}
Supervised randomization introduces heterogeneity in treatment probabilities into a randomized controlled experiment. We discuss the proposed approach as an extension to A/B testing through the introduction of a targeting model in several steps. We describe the process for one treatment and one control group in the online context where customers arrive in sequence, but note that the same process extends to more than one treatment group and other static settings. \\
A/B testing for treatment evaluation is an instance of randomized controlled experiments with a single treatment. Each arriving customer is randomly assigned to the treatment or control group. The probability to receive treatment is identical for all customers $e(x)=e$ with the probability for assignment to the control group  $1-e$. The probability of treatment assignment can be equal $e=1-e=0.5$ or imbalanced towards the preferred strategy for $e \in (0;1)$. 
As discussed in the literature review, imbalanced probabilities are used to control the costs of the experiment in practice. For the case of multiple treatments, a different probability can be assigned to each treatment. 

\begin{figure}[t]
  \begin{subfigure}[h]{0.48\textwidth}
  \centering
  \captionsetup{justification=centering}
    \includegraphics[width=\textwidth]{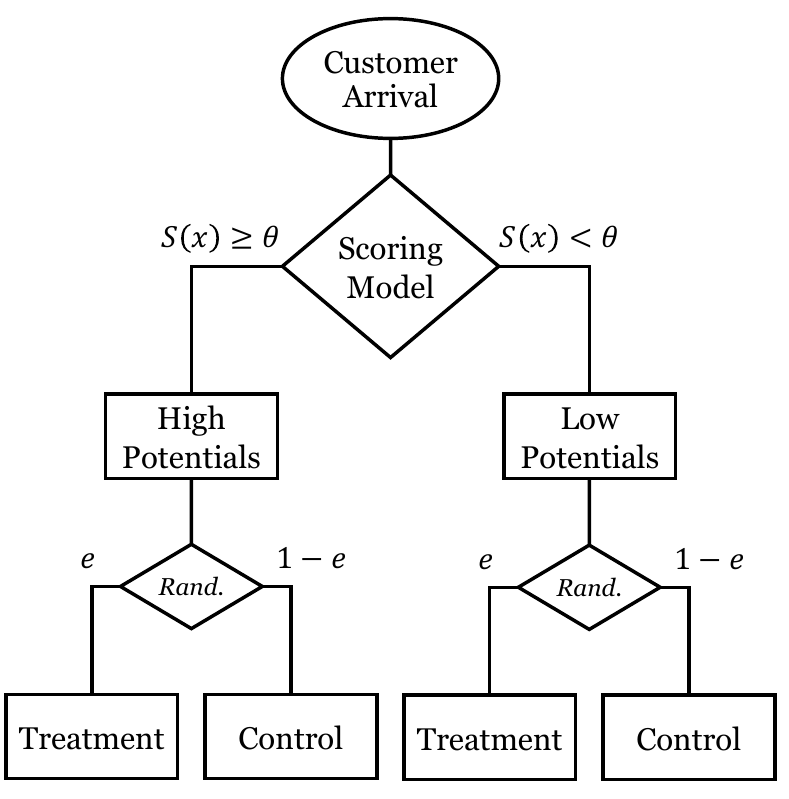}
    \caption{Full randomization (A/B test) \\ showing targeting policy}
    \label{fig:process_policy_AB}
  \end{subfigure}
\hspace{1em}%
     \begin{subfigure}[h]{0.48\textwidth}
  \centering
  \captionsetup{justification=centering}
    \includegraphics[width=\textwidth]{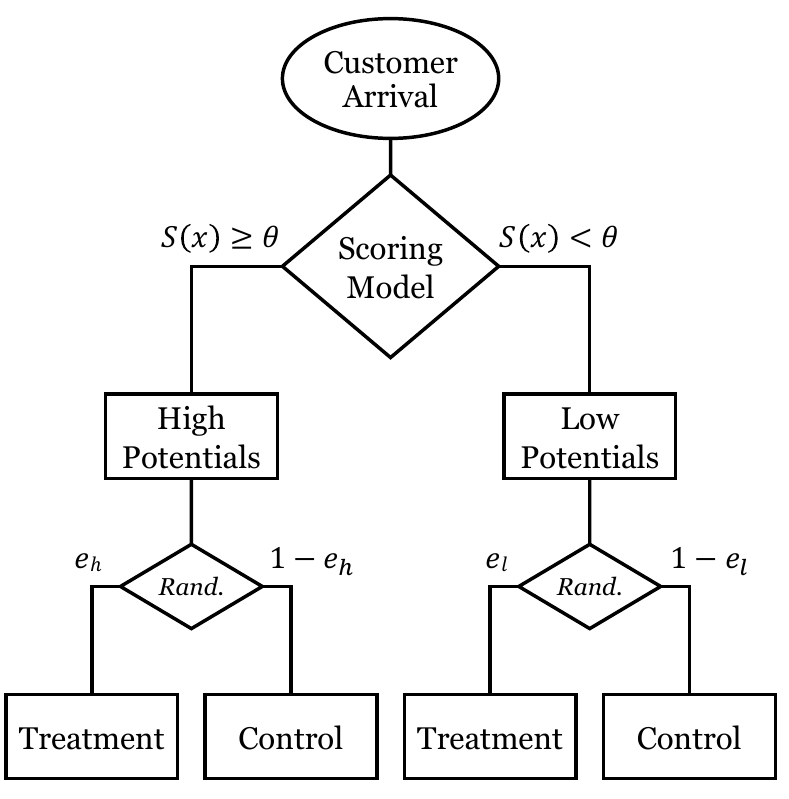}
    \caption{Supervised randomization \\ (two groups)}
    \label{fig:process_supervised_groups}
  \end{subfigure}
    \caption{Experimental design of full randomization (left) and supervised randomization (right). Note the heterogeneity in treatment probability for supervised randomization. \textit{Rand.} indicates random assignment} 
  \label{fig:process_classic}
\end{figure}

During regular business operation, the existing scoring model assigns a score $S(x)$ to each customer, where $S(x)$ could be an estimate of the conversion probability or the ITE. The model score $S(x)$ is compared to a threshold $\theta$ to classify customers into groups, where the group \textit{high potentials} consists of the customers with the higher score, e.g., the highest probability to respond positively to the treatment. The \textit{high potential} group would be targeted during regular operation, while the \textit{low potential} group would receive no treatment. Figure \ref{fig:process_policy_AB} visualizes a scoring model during A/B testing. For the purpose of experimental data collection, the classification and deterministic targeting is replaced by random targeting. Independent of group assignment, each customer has an equal probability to receive treatment $e(x)=e$. 

The proposed process of supervised randomization (Algorithm \ref{algo:supervised_randomization}) integrates the scoring model into the randomized treatment assignment. As an intermediate step, let the treatment probability be dependent on the classification by the targeting strategy as depicted in Figure \ref{fig:process_supervised_groups}. Different to the A/B test described in Figure \ref{fig:process_policy_AB}, where the targeting policy does not affect the treatment assignment, we now treat high potential customers with probability $e_h$ and low potential customers with probability $e_l$, where $e_h \neq e_l$ and $e_l,e_h \in (0;1)$. Note that $e_h$ and $e_l$ do not need to sum up to $1$. We increase the treatment probability in the high potential group relative to the low potential group by choosing $e_h$ and $e_l$ so that $e_h>e_l$. Thereby, more high potential customers than low potential customers are treated, in accordance with the scoring model and approximating the regular targeting policy. Simultaneously, we preserve a degree of randomization in the treatment/control assignment, since each customer has some probability to be assigned to the treatment or control group, respectively. The randomization is required to fulfill the overlap assumption (Eq. \ref{eq:overlap}) and should be large enough in practice to ensure coverage over the range of customer characteristics in both the treatment and control group. By violating the overlap assumption and setting $e_h=1$ and $e_l=0$, we recover the deterministic targeting policy of the classification model, where only customers in the \textit{high potential} group are treated. Note that if we choose a constant treatment probability $e_h = e_l $, the process simplifies to an A/B test on the whole population as shown in Figure \ref{fig:process_policy_AB}. 

We can further approximate individualized targeting by introducing more groups, each with a unique treatment probability $e_k$. Define a set of thresholds $[\theta_1, \theta_2, \ldots, \theta_K]$ and corresponding treatment probabilities $[e_1, e_2, \ldots, e_K]$ to target customer $i$ with probability $e_k$ for which $\theta_{k-1} < S(x_i) \leq \theta_{k}$. As above, we require $e_k \in (0;1)$ and $\sum_k e_k = 1$. By increasing the number of thresholds $K$, we approximate a continuous mapping $M: S(x) \rightarrow e$, where each customer is assigned an individual treatment probability $e_i$ proportional to the individual model score. \\
The specific mapping from model scores to treatment probability should follow the requirements of the application. We propose to determine the mapping by defining a set of $k$ equal-sized intervals on the range of the model score in the training data $[\min(S(X_{train})), \max(S(X_{train}))]$ and assigning a linearly increasing treatment probability $e_k$ to each interval, while setting the lowest treatment probability at $e_1=0.05$ and the highest at $e_K=0.95$. Note that asymmetric mappings result in a controlled shift of average treatment probability. The design of the mapping thus allows the straightforward extension to imbalanced supervised randomization.
%
\begin{algorithm}[t]
\SetAlgoLined
\DontPrintSemicolon
\LinesNotNumbered
\KwIn{Scoring model $S(\cdot)$; Treatment probability mapping $M(\cdot)$}
\KwOut{Treatment probability $e_{i,k}$; Treatment assignment $D_i \in \{0,1,  \ldots, K\}$; Outcome $Y_i$  }
 \For{i = 1, \ldots, N}{
  Observe customer $X_i$ \;
  Calculate customer score $s_{i,k} = S(X_i)$ \;
  Set treatment probability $e_{i,k} = M(s_{i,k})$ \;
  \BlankLine
  Draw treatment $D_i \sim \text{Categorical}(e_{i,k})$
  \BlankLine
  \eIf{$D_i == 0$}{
      Do not treat individual $i$ \;
      Observe outcome $Y_i(0)$
  }{
  \For{$k$ in $1,\ldots,K$}{
  \If{$D_i == k$}{
     Treat individual $i$ with treatment $k$ \;
     Observe outcome $Y_i(k)$
                 }%
}%
}%
}%
 \caption{Supervised Randomization for a Controlled Experiment with $K$ Treatments}
\label{algo:supervised_randomization}
\end{algorithm}
%
We reiterate that supervised randomization (Algorithm \ref{algo:supervised_randomization}) randomly assigns each customer to the treatment or control group, but adjusts the probability of this assignment based on the output of a scoring model, so that customers with higher score are treated with higher probability. The assigned individual treatment probabilities are logged and used in the subsequent analyses.

\subsection{Inverse probability weighting}
Using the targeting model to adjust the individual probability to receive treatment introduces a sampling bias into the experiment. The sampling bias is a direct result from the violation of independence between treatment probability and the individual characteristics via the scoring model. This type of selection bias commonly occurs in observational studies, where customers self-select into the treatment group, or in natural experiments. In both situations, the sample shows measurable distributional differences between the control and treatment group. Subsequent evaluation or model building need to correct for the selection bias to ensure unbiased estimates of the treatment effect. We will discuss IPW as a method that is easily integrated into model building and evaluation and discuss the doubly robust estimator as a recent extension. For a comprehensive overview of approaches including IPW see \cite{knaus2018machine}. The idea underlying all approaches is to weight each observation in the treatment or control group by the inverse of its respective probability to be assigned to the observed group. 

In contrast to observational studies where the treatment probability is estimated, the true probability at which customers receive the treatment is assigned actively based on a scoring model and a set of observed variables and is consequently known exactly under supervised randomization. Without the need to estimate the treatment probability from the data, we avoid confoundedness due to unobserved variables or misspecification of the propensity model by design. 

IPW restores the hypothetical distribution as it would look like in a fully randomized experiments by weighting every customer with regard to the individual treatment probability. Intuitively, customers who were assigned by chance to the treatment group, even though their characteristics result in a low treatment probability, are underrepresented in the treatment group. IPW assigns these customers a higher weight. For example, if the probability of being in the treatment group for a customer is $e(x) = 0.2$ then the observed outcome if this customer received treatment is multiplied by $1/e(x) = 1/0.2 =  5$. Vice versa, if the same customer was assigned to the control group, which happened with a probability of $1-e(x)) = 0.8$, the customer's outcome in the control group is weighted by $1/0.8 = 1.25$.\\
The IPW corrected ATE can then be estimated as: 

\begin{align}
    \widehat{ATE}_{IPW} = \frac{1}{N}\left(\sum_{i=1}^N \frac{D_iY_i}{e(X_i)} - \sum_{i=1}^N \frac{(1-D_i)Y_i}{1-e(X_i)}\right). \label{eq:IPW1}
\end{align}

In observational studies, the propensity scores are unknown and need to be estimated from observed covariates. The \textit{doubly robust} (DR) estimator is consistent and unbiased if only one of the models, the regression or the propensity score, is correctly specified \cite{lunceford2004stratification}:
\begin{align}
    \widehat{ATE}_{DR} = \frac{1}{N} \sum_{i=1}^{n} \frac{D_{i} Y_{i}-\left(D_{i}-e(X_i)\right) g_{1}\left(X_{i}\right)}{e(X_i)}-\frac{1}{N} \sum_{i=1}^{n} \frac{\left(1-D_{i}\right) Y_{i}+\left(D_{i}-e(X_i)\right) g_{0}\left(X_{i} \right)}{1-e(X_i)}.
    \label{DR}
\end{align}
Here $g_{D}\left(X_{i}\right)$ = $\operatorname{E}(Y|D,X_i=x)$ are models of the outcome variable on $x$, estimated separately for ${D \in \{0,1\}}$. 

Adjusting for the propensity score under full randomization has no effect on the point estimate for the average treatment effect. However, there is some evidence that even in fully randomized experiments the large-sample variance of the estimate can be reduced by using estimated propensity scores to control for random imbalance in covariates as well as orthogonalization with the mean as in the doubly robust \cite{williamson2014variance}. 

\section{Empirical Evaluation}\label{Section:Empirical}

We evaluate the proposed randomization procedures through a simulation study designed to represent a direct marketing setting, which is a common application of uplift modeling in marketing (see \cite{radcliffe2007using,devriendt2018literature}).\footnote{The \texttt{R} code for the empirical evaluation is available at \url{https://github.com/Humboldt-WI/supervised_randomization}.}

Since IPW correction with the true propensity score is feasible, ATE and ITE estimates are consistent under supervised randomization. The goal of the empirical study is to compare the increased conversion rate and cost savings due to supervised randomization with the loss in efficiency due to a less balanced sample. The efficiency of each randomization procedure has two dimensions, that is, 1) monetary cost of the experiment and 2) the quality of models trained on the data collected during the experiment measured on downstream tasks.\\
First, the campaign profit during the experiment provides a metric on which to compare the opportunity cost of different experimental designs. We compare the campaign profit under supervised randomization to the baseline of full randomization, which provides optimal data quality, and expect opportunity costs to be lower under the proposed supervised randomization. \\
Second, we evaluate the data generated from the experiment by comparing the predictive performance of ITE estimators trained on data under supervised randomization to the same estimators trained on data under full randomization. Our metrics of model performance are the mean absolute error to the true treatment effect (MAE), which is known in this simulation study but unknown in real-world settings, and the Qini coefficient, which is a standard metric in the uplift literature. The Qini coefficient is a rank metric similar to model lift based on the group-wise difference in conversion rates for customers ranked by their estimated treatment effect \cite{radcliffe2007using}.

\subsection{Simulation design}

 We compare the ATE and ITE estimates on experimental data collected under full and supervised randomization. An online evaluation of randomization procedures is challenging since it requires running a randomized experiment for each experimental design. We therefore evaluate the supervised randomization design in an offline study and leave online testing for future research. Our empirical Monte Carlo study uses real data to the extent possible to ensure a realistic setting in which we simulate the treatment effect and have full control over the treatment assignment \cite{nie2017QuasiOracleEstimationHeterogeneous, knaus2018machine}. The UCI Bank Marketing dataset \cite{moro2014DatadrivenApproachPredict} provides data on 45,211 customers of a Portuguese bank through 17 continuous or categorical variables covering individual socio-demographic and financial information, campaign details and macroeconomic indicators. All customers were subject to a phone marketing campaign promoting a term deposit and the target variable indicates if a customer has agreed to a deposit following the campaign.

Based on the available data, we simulate the individual treatment effect and hypothetical outcomes following the procedure of \cite{nie2017QuasiOracleEstimationHeterogeneous}. 
The treatment effect in real data can be a complex, non-linear function of a subset of observed variables and unobserved variables \cite{farrell2018DeepNeuralNetworks}. Therefore, we simulate the treatment effect as a combination of the twelve variables containing personal or macroeconomic information. 
The treatment effect as a non-linear combination of covariates is then modelled by a neural network of one hidden layer with the number of nodes equal to the number of input variables and sigmoid activation, initialized with random weights drawn from a standard Gaussian. To simulate the existence of unobserved covariates, e.g. due to privacy concerns, we remove variables with personal information on the customers' age and marital status from the subsequent analysis. \\
In marketing settings, we further expect the ATE to be positive but small and the ITE to be mostly non-negative as marketing theory suggests a direct marketing campaign to increase overall conversion, with potentially zero but rarely negative impact on customers \cite{hitsch2018heterogeneous}. We center the simulated ITE distribution at an ATE of 5\% and scale the standard deviation to 0.04 for 89\% of simulated ITE to be positive. For our application, an ATE of 5\% implies that the telephone campaign will convince an additional 5\% of randomly targeted customer to register a term deposit. 
Because all customers in the observed data have received the marketing treatment, we simulate the potential outcome without treatment by flipping outcome labels for observations chosen randomly in proportion to their treatment effect as in \cite{nie2017QuasiOracleEstimationHeterogeneous}.

Supervised randomization integrates an existing customer scoring model into the experimental design. A more accurate estimate from the existing model increases the extent to which potential cost savings are realized during experimentation. Noisier estimates of the scoring model lead to treatment assignment that is less profitable but has less influence on downstream tasks. In particular, supervised randomization with a noisy scoring model samples more evenly in the covariate space, thus mitigating the efficiency loss in downstream tasks, assuming a stochastic estimation error.
We control the quality of the existing targeting model by simulating a noisy causal model with predictions $\hat{\tau_i}=\tau_i + \varepsilon$ and $\varepsilon \sim \mathcal{N}(0,\sigma)$. We report results for $\sigma$=0.025, such that customers with ITE equal to the ATE have a 95\% chance to receive a prediction in the range [0,0.1] 
between the true and predicted treatment effect to provide a conservative estimate of the cost savings from supervised randomization. 
We split the data into four folds for cross validation and randomly assign treatment to each observation in the training data according to full or supervised randomization. For ITE estimation, we then estimate the ITE model on the training data and evaluate its prediction on the holdout fold. Since the random treatment assignment introduces additional randomness into the evaluation, we repeat the treatment assignment 50 times for each holdout fold and report the average over a total of 200 repetitions. 

\subsection{Statistical model performance analysis}

We first establish the effectiveness of supervised randomization independent of any application-specific cost setting. We evaluate the cost efficiency during experimentation through a comparison of conversion rates, ATE estimates based on their variance and ITE estimates based on uplift-specific performance metrics.

For cost efficiency, Table \ref{tab:experiment_summary} reports the mean target fraction and conversion rate for full randomization at equal probability, full but imbalanced randomization with treatment probability 0.66 and the proposed supervised randomization procedure. We provide statistics on targeting no or all customers for context. However, targeting no or all customers and other non-randomized targeting strategies do not allow experimental data collection. In other words, settings \textit{None} and \textit{All} are inapplicable in practice for targeting policy evaluation or treatment effect estimation. 
%
{
\renewcommand{\arraystretch}{1.2}

\begin{table}[ht]
\centering
\caption{Ratio of targeted customers and corresponding conversion rate under each randomization procedure. \textit{None}/\textit{All} denote targeting no/all customers for reference}
\label{tab:experiment_summary}
\begin{tabular}{lrr>{\em}rrr}
\toprule         
 & None & Full & Supervised & Full (Imb.) & All \\ 
  \midrule
Targeted Fraction of Customers & 0.000 & 0.500 & 0.500 & 0.666 & 1.000 \\ 
  Conversion Rate & 0.109 & 0.135 & 0.143 & 0.143 & 0.160 \\ 
\bottomrule
\end{tabular}
\end{table}
}
The target fractions for full randomization is 0.5 by definition and for imbalanced full randomization 0.66 by design. Small deviations from the target fraction are possible since treatment assignment is randomized on the individual level. 
The direction and ratio of the imbalance between the size of treatment to control group are in practice set by the experimenter to match the expected average treatment effect or marketing requirements, e.g. campaign budget. We chose a ratio of 2:1 in favor of targeting a larger group of customers following the most common design observed for customer targeting data in related research (see Table \ref{tab:uplift}). 

The conversion rate under each randomization provides an indirect measure of the campaign success with a higher conversion rate as an indicator of monetary returns. The increase in conversion rate from targeting no customers at 10.9\% to targeting all customers at 16\% reflects the simulated positive average treatment effect, specifically that customers are on average 5 percentage points more likely to convert after receiving the marketing treatment. 
During a fully randomized experiment, we observe an increase in the overall conversion rate by 2.6 percentage points to 13.5\%. At the same fraction of customers targeted, the proposed supervised randomization increases the conversion rate by another 0.8 points to 14.3\%. The improvement due to supervised randomization is the direct result of adjusting each customer's probability to be treated based on the targeting model and targeting customers with a high predicted treatment effect.\\ 
The benchmark strategy, imbalanced full randomization, increases treatment probability for all customers indiscriminately. The increase in individual treatment probability results in a conversion rate increase by 0.8 percentage points, identical to the increase under full randomization, but at a higher fraction of customers targeted.
The managerial implication is that supervised randomization achieves the same conversion rate as current best practice, while reducing the targeting rate with its associated costs by 24\%.

\begin{figure}[ht]
  \centering
    \includegraphics[width=0.7\textwidth]{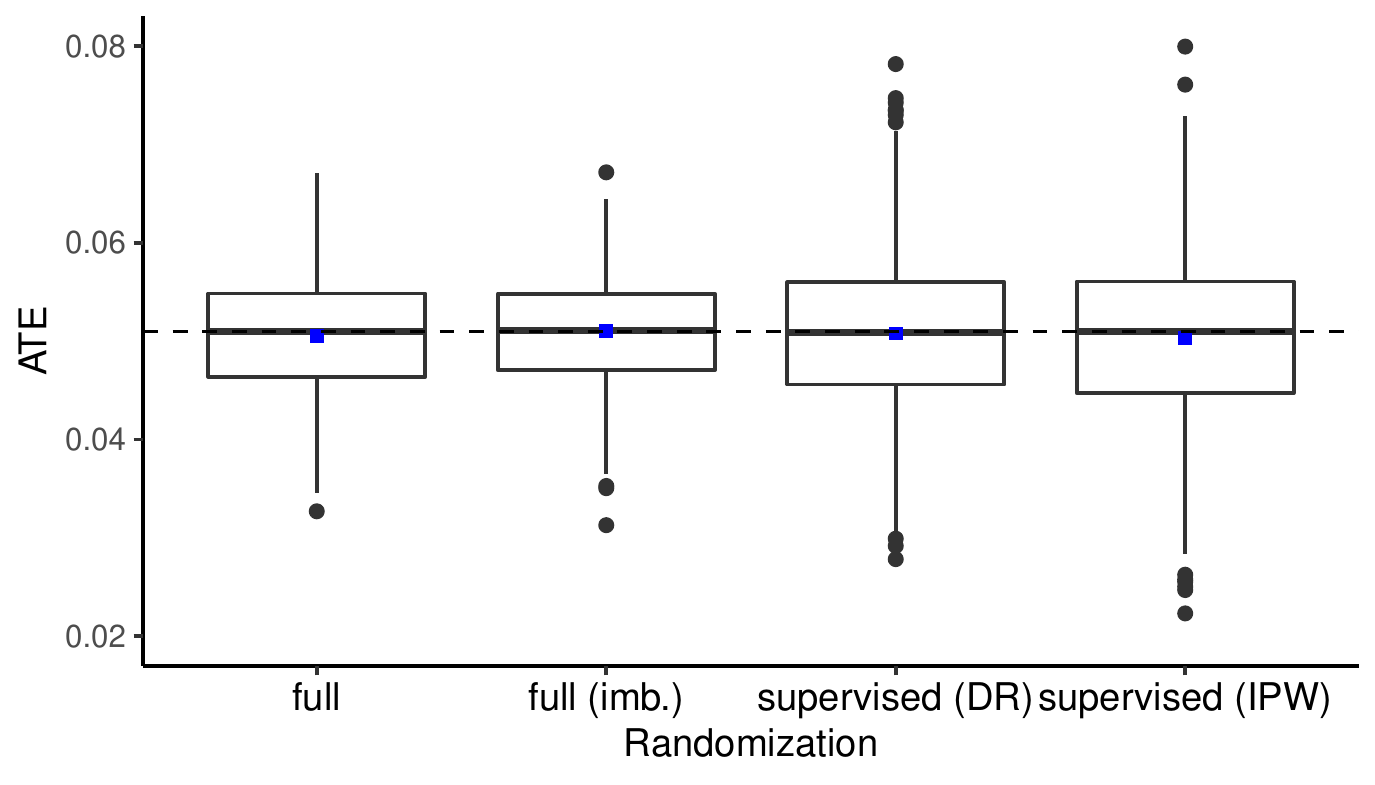}
  \caption{Estimated average treatment effect averaged over 200 iterations for each randomization procedure. The dashed horizontal line denotes the simulated true average treatment effect, dots within each boxplot denote the mean estimated ATE} 
  \label{fig:ATE}
\end{figure}

The higher conversion rate from targeting randomization towards relevant customers comes at the downside of collecting less data for customer groups with very high or very low treatment probability. While we can use the logged treatment probabilities to optimally correct for the sampling bias that is introduced by supervised randomization, estimates of the treatment effect will exhibit higher uncertainty through higher variance. Figure \ref{fig:ATE} shows the estimated ATE under each randomization procedure. We see that 1) deviations from full randomization in the form of imbalanced full randomization and supervised randomization return unbiased estimates and 2) the overall variance from the true value and the number of extreme deviations increases when moving from full randomization to supervised randomization. \\
 A Kruskal-Wallis test verifies that there is no significant difference in the mean point estimate among the four settings (\textit{df}=3, $\chi^2$ = 1.00). We are thus able to verify the theoretical exposition and to show that the selection bias introduced by supervised randomization can be corrected for by applying either IPW or DR as described above. The additional uncertainty due to supervised randomization is less pronounced when using DR to correct for heterogeneous treatment probabilities instead of IPW. DR estimates exhibit a significantly lower variance when compared to IPW estimates, based on a Levene-test for homogeneity of variance (\textit{df}=1, F=10.29).   

As uplift applications are concerned with the estimation of individualized treatment effects for customer scoring, we proceed to evaluate the model performance of two causal models on the data collected under each randomization procedure. We select the two-model approach using logistic regression and the causal forest and report the performance of using the ATE as a constant prediction for reference. Since our focus is on the comparison of the randomization procedures rather than a comparison of ITE estimators, we manually set the parameters for the causal forest as follows: number of variables tried at each split (mtry) = 7, number of trees = 500, minimum node size = 20 and sample fraction for honest tree building = 0.5. Model predictions are evaluated using the mean absolute error to the true ITE and the Qini score on holdout data.

{
\renewcommand{\arraystretch}{1.2}

\begin{table}[ht!]
\centering
\caption{Average profit-agnostic performance of causal models for each randomization procedure. We evaluate the models using the MAE to the (simulated) true treatment effect (lower is better) and the Qini coefficient (higher is better)} 
\label{tab:ITE_KPI}
\begin{tabular}{lrrrrrr} 
\toprule
              & \multicolumn{2}{c}{ATE} & \multicolumn{2}{c}{Two-Model (LR)} & \multicolumn{2}{c}{Causal Forest} \\ 
              \cmidrule(l{2pt}r{2pt}){2-3} \cmidrule(l{2pt}r{2pt}){4-5} \cmidrule(l{2pt}r{2pt}){6-7}
              & MAE        & Qini       & MAE              & Qini            & MAE             & Qini            \\ \midrule
Full        & 0.0324 & -& 0.0353 & 0.0045 & 0.0276 & 0.0056 \\ 
Full (imb.) & 0.0324 & - & 0.0357 & 0.0045 & 0.0275 & 0.0057 \\ 
\textit{Supervised} & \textit{0.0325} & \textit{-} & \textit{0.0383} & \textit{0.0041} & \textit{0.0295} & \textit{0.0047} \\ 
\bottomrule
\end{tabular}
\end{table}
}

We identify two takeaways in Table \ref{tab:ITE_KPI}. First, the causal forest outperforms the two-learner approach on both MAE and Qini. The performance difference is consistent over all randomization procedures with the causal forest resulting in a MAE lower by about 0.008 points and a Qini higher by 0.001 points. Second, we observe that deviating from full randomization to supervised randomization leads to the expected decrease in model performance. Under supervised randomization, the difference to full randomization for the two-model approach is 0.003 points MAE and 0.0004 points Qini and for the causal forest 0.002 points MAE and 0.001 points Qini. Imbalanced full randomization at $e=0.66$ shows no substantial performance decrease compared to balanced full randomization, although additional experiments indicate lower performance at higher levels of imbalance. The subsequent profit-based analysis aims to provide a comprehensible evaluation of the observed differences in a business context.

\subsection{Profit analysis}

We proceed to empirically show the extent to which supervised randomization can reduce the cost of running a randomized experiment and the size of the expected trade-off measured by the performance of models trained on the collected data. 
The profit setting for telephone marketing is described by the gross profit resulting from a conversion and the variable contact cost of making a call to the customer. 
If we assume a constant interest margin for the bank, the gross profit from a one-year term deposit $\Omega$ is equivalent to the net interest margin $m$ and the deposit amount $A$, $\Omega_i = m A_i$. \\
The incremental gross profit due to a marketing campaign is defined as change in the conversion probability, the treatment effect, to earn the gross profit on conversion minus the contact cost $c$, i.e. $\Delta \Omega_i = \tau_i  m A_i - c$.\\
Given an accurate estimate of the treatment effect $\tau_i$, the decision to target a specific customer is profitable when the predicted incremental gross profit for the customer is positive, i.e. $\hat{\tau}_i  m A_i - c > 0$.

To simplify interpretation, we consider cost ratios in the range of $[5,10,\ldots,50]$ to 1. Evaluation over a range of cost settings ensures the robustness of our results and allows generalization to a variety of profit and cost scenarios that may arise across banks or industries, e.g. for catalog marketing. We can empirically confirm the plausibility of the range of cost ratios by analyzing the ratio of customers which are targeted under each cost setting. For cost ratios below 10:1 and above 50:1, individual targeting policies are dominated by indiscriminate targeting of no or all customers, respectively.
The cost ratio corresponds to different values of the interest margin $m$ and deposit amount $A$ at standardized contact cost. Assuming a constant amount of the term deposit $\overline{A}$ for each customer, the cost ratio can be interpreted as the ratio between the gross profit over a range of interest margins $m$ standardized to contact costs of $c=1$ per contact. 

We evaluate the cost-saving potential of using supervised randomization during experimentation based on the campaign profit resulting from a randomized experiment for each randomization procedure. 
We report the campaign profit per prospective customer and the difference in campaign profit relative to full randomization in Table \ref{tab:experiment_profit}. As above, we include targeting no customers and targeting all customers for reference, but stress that non-randomized targeting strategies do not allow experimental data collection, making them inapplicable for causal modeling in practice.
{
\renewcommand{\arraystretch}{1.2}

\begin{table}[ht!]
\centering
\caption{Campaign profit (per customer) for randomized experiments under each randomization procedure and across purchase margins. \textit{None}/\textit{All} denote targeting no/all customers for reference. \textit{Full (Imb.)} denotes full randomization with a treatment probability of 66\%} 
\label{tab:experiment_profit}
\begin{tabular}{crr>{\em}rrr}
\toprule
 Conversion & \multicolumn{5}{c}{Campaign profit per customer (\EUR{})}  \\
  \cline{2-6} 
 Value (\EUR{})   & None   & Full   & Supervised & Full (Imb.)  & All              \\
 \midrule
  10 & 1.09 & 0.85 & 0.93 & 0.76 & 0.60 \\
  15 & 1.64 & 1.52 & 1.65 & 1.48 & 1.40 \\
  20 & 2.18 & 2.19 & 2.37 & 2.19 & 2.20 \\
  25 & 2.73 & 2.86 & 3.09 & 2.91 & 3.00 \\
  30 & 3.27 & 3.54 & 3.80 & 3.62 & 3.80 \\
  35 & 3.82 & 4.21 & 4.52 & 4.34 & 4.60 \\
  40 & 4.36 & 4.88 & 5.24 & 5.05 & 5.40 \\
  45 & 4.91 & 5.56 & 5.96 & 5.77 & 6.20 \\
  50 & 5.45 & 6.23 & 6.67 & 6.48 & 7.00 \\
\bottomrule
\end{tabular}
\end{table}
}

The empirical results in Table \ref{tab:experiment_profit} support the proposition that supervised randomization increases the campaign profit during experimentation relative to full randomization for the full range of conversion values we consider in this study. 
 In relative terms, supervised randomization increases the experimental campaign profit by 7.1-9.4\% compared to full randomization and by 2.9-8.2\% compared to imbalanced randomization.
 
For a conversion value of \EUR{10}, we observe a marginal profit of \EUR{0.85} per customer under full randomization and a marginal profit of \EUR{0.93} under supervised randomization. The absolute increase in campaign profit is more pronounced when the cost ratio is higher. A value of \EUR{50} corresponds to a marginal profit per customer of \EUR{6.23} under full randomization compared to \EUR{6.67} under the proposed supervised randomization. Cost savings per customer compared to full randomization amount to \EUR{0.08} and \EUR{0.44}, respectively.\\
We translate the per customer savings to an experimental campaign of 40,000 prospective customers, who are randomly targeted. This is the size of the observed telephone marketing campaign and, with less observations than 9 of the 11 experimental marketing campaigns summarized in Table \ref{tab:uplift}, may provide a conservative estimate. The total cost savings per experiment when replacing full randomization with supervised randomization translate to \EUR{3,200} for a marginal profit of \EUR{10}, \EUR{10,400} for a marginal profit of \EUR{30} and \EUR{17,600} for a marginal profit of \EUR{50}. Experiment costs and the related savings arise whenever data is collected for policy evaluation or (re-)estimation of the customer scoring model.

For conversion values greater or equal \EUR{20}, targeting all customers is more profitable than not targeting any customer. The imbalanced full randomization, which we identify as standard in practice, is more profitable than full randomization only at values above \EUR{20}. At these values, imbalanced randomization achieves savings of 0 to \EUR{0.25} per customer compared to full randomization for conversion values between \EUR{20} and \EUR{50}, respectively. Compared to imbalanced full randomization, the proposed supervised randomization generates additional cost savings per customer of about \EUR{0.18} for all values between \EUR{20} and \EUR{50}. 
Again translated to an experiment campaign of 40,000 prospective customers, the total cost savings per experiment of supervised randomization when compared to the industry-standard range from \EUR{7,200} for a marginal profit of \EUR{20} to \EUR{7,600} for a marginal profit of \EUR{50}. Note that it is possible to combine supervised randomization with imbalanced targeting. Increasing the average treatment probability through a custom treatment probability mapping may further increase campaign profit in settings where treatment is highly profitable. 

Having discussed the expected cost savings during experimentation, we next discuss the opportunity costs on downstream tasks associated with the increase in model uncertainty under supervised randomization. We first report the campaign profit per customer when customers are targeted by the two-model approach or causal forest and each model is trained on experimental data collected under the different randomization procedures. 

{
\renewcommand{\arraystretch}{1.2}

\begin{table}[ht]
\centering
\caption{Campaign profit using targeting models trained on data collected under each randomization procedure. We evaluate the campaign profit per customer over a range of cost ratios} 
\label{tab:ITE_profit}
\begin{tabular}{crr>{\em}rrr>{\em}r}
\toprule
Conversion & \multicolumn{3}{c}{Two-Model (Logit)} & \multicolumn{3}{c}{Causal Forest}   \\
\cmidrule(l{2pt}r{2pt}){2-4}\cmidrule(l{2pt}r{2pt}){5-7}
  Value (\EUR{})               & Simple & Simple (Imb.) & Supervised    & Simple & Simple (Imb.) & Supervised \\

\midrule
  10 & 1.06 & 1.06 & 1.06 & 1.09 & 1.09 & 1.09 \\ 
  15 & 1.65 & 1.65 & 1.65 & 1.66 & 1.67 & 1.65 \\ 
  20 & 2.33 & 2.33 & 2.32 & 2.36 & 2.36 & 2.33 \\ 
  25 & 3.07 & 3.06 & 3.05 & 3.11 & 3.12 & 3.08 \\ 
  30 & 3.83 & 3.82 & 3.80 & 3.89 & 3.89 & 3.84 \\ 
  35 & 4.60 & 4.60 & 4.57 & 4.67 & 4.67 & 4.62 \\ 
  40 & 5.38 & 5.38 & 5.35 & 5.45 & 5.45 & 5.41 \\ 
  45 & 6.16 & 6.16 & 6.13 & 6.24 & 6.24 & 6.20 \\ 
  50 & 6.95 & 6.95 & 6.91 & 7.03 & 7.03 & 7.00 \\ 
\bottomrule
\end{tabular}
\end{table}
}

Table \ref{tab:ITE_profit} shows that the expected decrease in profit for scoring models trained on data collected under supervised randomization is small but observable in the order of 1\% of the absolute campaign profit per customer. 
For a basket margin of \EUR{30}, the two-model logistic regressions achieve a campaign profit of \EUR{3.83} per customer under full randomization and a campaign profit of \EUR{3.80} under supervised randomization, a decrease of 0.8\%. The causal forest achieves a campaign profit of \EUR{3.89} when trained on data from experiments under full randomization with a decrease by 1.3\% to \EUR{3.84} under supervised randomization. Compared over all values, supervised randomization induces a decrease in per customer profit between \EUR{0} and \EUR{0.04} for the two-model approach and \EUR{0} and \EUR{0.05} for the causal forest compared to full randomization. 

\section{Conclusion}

Customer targeting is a continuously growing and widely studied application of scoring models. While research has focused on the prediction of future customer behavior to inform decision-making, a growing research stream has established uplift models to estimate the causal effect of a marketing action on each customer based on observed customer characteristics. The training and evaluation of causal models require data collected through experiments, in which customers are randomly assigned to treatments. However, experimental data collection incurs high costs by temporarily replacing an established targeting policy with random targeting. 

We propose supervised randomization as a solution to reduce the cost of experimentation by integrating an existing scoring model into the experimental design. By mapping model scores to individual treatment propensities, we are able to target more profitable customers while maintaining stochastic treatment assignment. An empirical Monte Carlo study on telemarketing shows that supervised targeting can reduce the cost of an experimental campaign on 40,000 prospective customers by 7.1-9.4\% compared to full randomization and 2.9-8.2\% compared to imbalanced randomization, depending on the specific profit-cost ratio. 

Active management of treatment assignment during experimentation leads to an overrepresentation of profitable customers in the treatment group, which causes selection bias when standard estimators are applied to estimate treatment effects. We consequently summarize inverse probability weighting and doubly robust estimation as well-studied methods to correct for selection bias when estimating average and individualized treatment effects. We show that the estimated treatment effects are unbiased and provide indicators of the increase in uncertainty related to supervised randomization. Empirical evaluation indicates that higher uncertainty of the scoring model may lead to a decrease in campaign profit by 0.8-1.3\% depending on the specific profit-cost ratio. Further evaluation in real-world experiments is necessary to establish net cost savings in practice. 

Overall, we argue that the methodology developed in the medical and econometric literature has not yet been fully studied and applied in the uplift setting. Doubly robust estimation serves as one example of a wider set of tools to correct for selection issues in the data. We further identify experimental data collection as a fundamental part of causal modeling. We expect that supervised randomization provides a first step towards a wider analysis of practical experimental design.


\bibliographystyle{unsrt}
\bibliography{references.bib}

\begin{thebibliography}{10}

\bibitem{statista2019ecommerce}
Statista.
\newblock ecommerce.
\newblock \url{https://www.statista.com/outlook/243/ecommerce}, 2019.

\bibitem{statista2017advertising}
Statista.
\newblock Advertising spending in the catalog, mail-order houses industry in
  the united states.
\newblock
  \url{https://www.statista.com/statistics/470620/catalog-mail-order-houses-industry-ad-spend-usa/},
  2017.

\bibitem{olson2012direct}
David~L. Olson and Bongsug Chae.
\newblock Direct marketing decision support through predictive customer
  response modeling.
\newblock {\em Decision Support Systems}, 54(1):443--451, 2012.

\bibitem{gubela2017revenue}
Robin~Marco Gubela, Stefan Lessmann, Johannes Haupt, Annika Baumann, Tillmann
  Radmer, and Fabian Gebert.
\newblock Revenue uplift modeling.
\newblock In {\em Proceedings of the 38th International Conference on
  Information Systems (ICIS’17)}, Seoul, South Korea, 2017.

\bibitem{devriendt2018literature}
Floris Devriendt, Darie Moldovan, and Wouter Verbeke.
\newblock A literature survey and experimental evaluation of the
  state-of-the-art in uplift modeling: A stepping stone toward the development
  of prescriptive analytics.
\newblock {\em Big Data}, 6(1):13--41, 2018.

\bibitem{radcliffe2007using}
Nicholas~J Radcliffe.
\newblock Using control groups to target on predicted lift: Building and
  assessing uplift models.
\newblock {\em Direct Marketing Journal, Direct Marketing Association Analytics
  Council}, 1:14--21, 2007.

\bibitem{ascarza2018RetentionFutilityTargeting}
Eva Ascarza.
\newblock Retention {{futility}}: {{targeting high risk customers might be
  ineffective}}.
\newblock {\em Journal of Marketing Research}, 55(1), 2018.

\bibitem{powers2018methods}
Scott Powers, Junyang Qian, Kenneth Jung, Alejandro Schuler, Nigam~H Shah,
  Trevor Hastie, and Robert Tibshirani.
\newblock Some methods for heterogeneous treatment effect estimation in high
  dimensions.
\newblock {\em Statistics in Medicine}, 37(11):1767--1787, 2018.

\bibitem{knaus2018machine}
Michael~C. Knaus, Michael Lechner, and Anthony Strittmatter.
\newblock Machine {{Learning Estimation}} of {{Heterogeneous Causal Effects}}:
  {{Empirical Monte Carlo Evidence}}.
\newblock {\em IZA Discussion Paper}, 12039, 2019-01.

\bibitem{rosenbaum1983central}
Paul~R Rosenbaum and Donald~B Rubin.
\newblock The central role of the propensity score in observational studies for
  causal effects.
\newblock {\em Biometrika}, 70(1):41--55, 1983.

\bibitem{Ascarza2017target}
Eva Ascarza, Peter Ebbes, Oded Netzer, and Matthew Danielson.
\newblock Beyond the target customer: social effects of customer relationship
  management campaigns.
\newblock {\em Journal of Marketing Research}, 54(3):347--363, 2017.

\bibitem{hitsch2018heterogeneous}
Günter~J Hitsch and Sanjog Misra.
\newblock Heterogeneous treatment effects and optimal targeting policy
  evaluation, 2018.

\bibitem{imbens2009RecentDevelopmentsEconometrics}
Guido~W Imbens and Jeffrey~M Wooldridge.
\newblock Recent developments in the econometrics of program evaluation.
\newblock {\em Journal of Economic Literature}, 47(1):5--86, 2009-03.

\bibitem{gordon2019ComparisonApproachesAdvertising}
Brett~R Gordon, Florian Zettelmeyer, Neha Bhargava, and Dan Chapsky.
\newblock A comparison of approaches to advertising measurement: Evidence from
  big field experiments at {{Facebook}}.
\newblock {\em Marketing Science}, 38(2):193--364, 2019.

\bibitem{kunzel2019MetalearnersEstimatingHeterogeneousa}
Sören~R Künzel, Jasjeet~S Sekhon, Peter~J Bickel, and Bin Yu.
\newblock Metalearners for estimating heterogeneous treatment effects using
  machine learning.
\newblock {\em Proceedings of the National Academy of Sciences},
  116(10):4156--4165, 2019.

\bibitem{gubela2019conversion}
Robin~Marco Gubela, Artem Bequé, Fabian Gebert, and Stefan Lessmann.
\newblock Conversion uplift in e-commerce: a systematic benchmark of modeling
  strategies.
\newblock {\em International Journal of Information Technology \& Decision
  Making}, 18(3):747--791, 2019.

\bibitem{farrell2018DeepNeuralNetworks}
Max~H. Farrell, Tengyuan Liang, and Sanjog Misra.
\newblock Deep {{Neural Networks}} for {{Estimation}} and {{Inference}}:
  {{Application}} to {{Causal Effects}} and {{Other Semiparametric Estimands}}.
\newblock {\em arXiv preprint arXiv:1809.09953}, 2018.

\bibitem{lo2002true}
Victor~SY Lo.
\newblock The true lift model: a novel data mining approach to response
  modeling in database marketing.
\newblock {\em ACM SIGKDD Explorations Newsletter}, 4(2):78--86, 2002.

\bibitem{zaniewicz2013support}
Lukasz Zaniewicz and Szymon Jaroszewicz.
\newblock Support {{Vector Machines}} for {{Uplift Modeling}}.
\newblock In {\em Proceedings of the 13th {{International Conference}} on
  {{Data Mining Workshops}}}, pages 131--138. {IEEE}, 2013.

\bibitem{rzepakowski2012decision}
Piotr Rzepakowski and Szymon Jaroszewicz.
\newblock Decision trees for uplift modeling with single and multiple
  treatments.
\newblock {\em Knowledge and Information Systems}, 32(2):303--327, 2012.

\bibitem{athey2016recursive}
Susan Athey and Guido Imbens.
\newblock Recursive partitioning for heterogeneous causal effects.
\newblock {\em Proceedings of the National Academy of Sciences},
  113(27):7353--7360, 2016.

\bibitem{athey2019GeneralizedRandomForests}
Susan Athey, Julie Tibshirani, and Stefan Wager.
\newblock Generalized random forests.
\newblock {\em The Annals of Statistics}, 47(2):1148--1178, 2019.

\bibitem{swaminathan2015BatchLearningLogged}
Adith Swaminathan and Thorsten Joachims.
\newblock Batch learning from logged bandit feedback through counterfactual
  risk minimization.
\newblock {\em Journal of Machine Learning Research}, 16:1731−1755, 2015.

\bibitem{athey2017EfficientPolicyLearning}
Susan Athey and Stefan Wager.
\newblock Efficient policy learning.
\newblock {\em arXiv preprint arXiv:1702.02896v2}, 2017.

\bibitem{kane2014mining}
Kathleen Kane, Victor~SY Lo, and Jane Zheng.
\newblock Mining for the truly responsive customers and prospects using
  true-lift modeling: Comparison of new and existing methods.
\newblock {\em Journal of Marketing Analytics}, 2(4):218--238, 2014.

\bibitem{hansotia2002direct}
Behram~J Hansotia and Bradley Rukstales.
\newblock Direct marketing for multichannel retailers: Issues, challenges and
  solutions.
\newblock {\em Journal of Database Marketing \& Customer Strategy Management},
  9(3):259--266, 2002.

\bibitem{guelman2014optimal}
Leo Guelman, Montserrat Guill{\'e}n, and Ana~Mar{\'i}a P{\'e}rez~Mar{\'i}n.
\newblock Optimal personalized treatment rules for marketing interventions: A
  review of methods, a new proposal, and an insurance case study.
\newblock {\em UB Riskcenter Working Paper Series}, 2014/06, 2014.

\bibitem{chickering2000decision}
David~Maxwell Chickering and David Heckerman.
\newblock A decision theoretic approach to targeted advertising.
\newblock In {\em Proceedings of the Sixteenth Conference on Uncertainty in
  Artificial Intelligence}, pages 82--88. Morgan Kaufmann Publishers Inc.,
  2000.

\bibitem{diemert2018large}
Eustache Diemert, Artem Betlei, Christophe Renaudin, and Massih-Reza Amini.
\newblock A large scale benchmark for uplift modeling.
\newblock In {\em Proceedings of the AdKDD and TargetAd Workshop, KDD}, London,
  United Kingdom, 2018. ACM.

\bibitem{guelman2015uplift}
Leo Guelman, Montserrat Guill{\'e}n, and Ana~M P{\'e}rez-Mar{\'i}n.
\newblock Uplift random forests.
\newblock {\em Cybernetics and Systems}, 46(3-4):230--248, 2015.

\bibitem{hillstrom2008mine}
Kevin Hillstrom.
\newblock The minethatdata e-mail analytics and data mining challenge.
\newblock MineThatData Blog.
  \url{https://blog.minethatdata.com/2008/03/minethatdata-e-mail-analytics-and-data.html},
  2008.

\bibitem{hansotia2002incremental}
Behram Hansotia and Brad Rukstales.
\newblock Incremental value modeling.
\newblock {\em Journal of Interactive Marketing}, 16(3):35--46, 2002.

\bibitem{schulz2002generation}
Kenneth~F. Schulz and David~A. Grimes.
\newblock Generation of allocation sequences in randomised trials: chance, not
  choice.
\newblock {\em The Lancet}, 359(9305):515--519, 2002.

\bibitem{lachin1988randomization}
John~M. Lachin, John~P. Matts, and L.~J. Wei.
\newblock Randomization in clinical trials: Conclusions and recommendations.
\newblock {\em Controlled Clinical Trials}, 9(4):365--374, 1988.

\bibitem{rosenberger1993use}
William~F. Rosenberger and John~M. Lachin.
\newblock The use of response-adaptive designs in clinical trials.
\newblock {\em Controlled Clinical Trials}, 14(6):471--484, 1993.

\bibitem{schwartz2017CustomerAcquisitionDisplay}
Eric~M. Schwartz, Eric~T. Bradlow, and Peter~S. Fader.
\newblock Customer acquisition via display advertising using multi-armed bandit
  experiments.
\newblock {\em Marketing Science}, 36(4):500--522, 2017-07.

\bibitem{horvitz1952generalization}
Daniel~G Horvitz and Donovan~J Thompson.
\newblock A generalization of sampling without replacement from a finite
  universe.
\newblock {\em Journal of the American Statistical Association},
  47(260):663--685, 1952.

\bibitem{robins1994estimation}
James~M Robins, Andrea Rotnitzky, and Lue~Ping Zhao.
\newblock Estimation of regression coefficients when some regressors are not
  always observed.
\newblock {\em Journal of the American Statistical Association},
  89(427):846--866, 1994.

\bibitem{caliendo2012cost}
Marco Caliendo, Michel Clement, Dominik Papies, and Sabine Scheel-Kopeinig.
\newblock The cost impact of spam filters: Measuring the effect of information
  system technologies in organizations.
\newblock {\em Information Systems Research}, 23(3):1068--1080, 2012.

\bibitem{schnabel2016recommendations}
Tobias Schnabel, Adith Swaminathan, Ashudeep Singh, Navin Chandak, and Thorsten
  Joachims.
\newblock Recommendations as treatments: Debiasing learning and evaluation.
\newblock In {\em Proceedings of the 33rd International Conference on Machine
  Learning}, volume~48, pages 1670--1679, 2016.

\bibitem{lunceford2004stratification}
Jared~K Lunceford and Marie Davidian.
\newblock Stratification and weighting via the propensity score in estimation
  of causal treatment effects: A comparative study.
\newblock {\em Statistics in Medicine}, 23(19):2937--2960, 2004.

\bibitem{williamson2014variance}
Elizabeth~J Williamson, Andrew Forbes, and Ian~R White.
\newblock Variance reduction in randomised trials by inverse probability
  weighting using the propensity score.
\newblock {\em Statistics in Medicine}, 33(5):721--737, 2014.

\bibitem{nie2017QuasiOracleEstimationHeterogeneous}
Xinkun Nie and Stefan Wager.
\newblock Quasi-{{Oracle Estimation}} of {{Heterogeneous Treatment Effects}}.
\newblock {\em arXiv preprint arXiv:1712.04912}, 2017.

\bibitem{moro2014DatadrivenApproachPredict}
Sérgio Moro, Paulo Cortez, and Paulo Rita.
\newblock A data-driven approach to predict the success of bank telemarketing.
\newblock {\em Decision Support Systems}, 62:22--31, 2014.

\end{thebibliography}

\end{document}